\documentclass[sigconf]{acmart}

\usepackage[T1]{fontenc}

\usepackage{booktabs}
\usepackage[utf8]{inputenc}
\usepackage{multirow}
\usepackage{subfig}
\usepackage{xspace}
\usepackage{flushend}
\usepackage{enumitem}

\acmConference[KDD'19  AutoML Workshop]{The Third International Workshop on Automation in Machine Learning}{August 2019}{Anchorage, AK} 
\acmYear{2019}

\begin{document}

\title[Automatic Historical Feature Generation]{Automatic Historical Feature Generation through Tree-based Method in Ads Prediction}

\author{Hongjian Wang}
\affiliation{%
  \institution{Twitter}
}
\email{hongjianw@twitter.com}
\author{Qi Li}
\affiliation{%
  \institution{Google}
}
\email{qilqil@google.com}
\author{Lanbo Zhang}
\affiliation{%
  \institution{Airbnb}
}
\email{lanbozhang@gmail.com}
\author{Yue Lu}
\affiliation{%
  \institution{Facebook}
}
\email{yuel@fb.com}
\author{Steven Yoo}
\affiliation{%
  \institution{Twitter}
}
\email{syoo@twitter.com}
\author{Srinivas Vadrevu}
\affiliation{%
  \institution{Twitter}
}
\email{svadrevu@twitter.com}
\author{Zhenhui Li}
\affiliation{%
  \institution{Penn State University}
}
\email{jessieli@ist.psu.edu}

\begin{abstract}
Historical features are important in ads click-through rate (CTR) prediction, because they account for past engagements between users and ads. In this paper, we study how to efficiently construct historical features through counting features. The key challenge of such problem lies in how to automatically identify counting keys. We propose a tree-based method for counting key selection. The intuition is that a decision tree naturally provides various combinations of features, which could be used as counting key candidate. In order to select personalized counting features, we train one decision tree model per user, and the counting keys are selected across different users with a frequency-based importance measure. To validate the effectiveness of proposed solution, we conduct large scale experiments on Twitter\footnote{Twitter and Tweets are registered marks of Twitter Inc.} video advertising data. In both online learning and offline training settings, the automatically identified counting features outperform the manually curated counting features.
\end{abstract}

\maketitle
\renewcommand{\shortauthors}{Wang et al.}

\section{Introduction}

The data rich advertising environment provides ample opportunities for machine learning to play a key role. Ads CTR prediction problem is one of the most important tasks. At Twitter, the ads are shown as the form of a Promoted Tweet. Twitter users usually consume Tweets in their home timeline, where the Tweets created by the people they follow are presented. Among the list of user generated Tweets, Promoted Tweets are injected in a set of given positions. The selection of which ads to show is based on the combination of predicted CTR and advertising bidding. We use impression refer to showing an ad to a user, and engagement refers to user acting on the ads, i.e. clicking on the link, watching the video ads, etc.

\emph{Contextual features} are the most straightforward features for CTR prediction, which account for the current context of an impression, including user information, ads content, nearby Tweets, etc. However, with contextual features alone it is very difficult to learn a CTR prediction model due to two data sparsity issues. First, the feature space is huge, such as ID-based feature. Second, the number of samples for each user-ad pair is small.

To address the issues above, \emph{historical features} are added to the prediction model. The historical features capture the past engagement information of the user-ad pair, while avoid explicit matching of user or ads. For example, past CTR is a strong feature in CTR model. Namely, even if user $u$ only saw ad $d$ once and engaged with it (a past CTR of $100\%$), the model is confident to predict a high CTR for showing $d$ to $u$ in the future. As a result, instead of maintaining the user ID and ad ID as contextual features, we could simply add one historical feature -- past CTR. Both contextual features and historical features are indispensable for CTR prediction~\cite{he2014practical}. However, it has shown that the historical features play a more important role than the contextual features~\cite{he2014practical}.

We use \emph{counting feature} refer to the counting-based implementation of historical features. Furthermore, we could count the past engagements under different conditions, which are called \emph{counting keys}. When selecting proper counting keys, there are a lot of choices. Basically, we can count on any categorical features, and even the combinations of them. In this paper, we aim to efficiently extract useful counting  keys. This task is challenging for the following three reasons. First, the exhaustive search is impossible, given an exponential search space. Second, the counting keys suffers from sparsity, i.e. the coverage of most counting keys are low. Third, as the size of counting keys increases, it is hard to semantically interpret the counting keys.

In this paper, we propose a method to automatically select counting keys from tree-based models. The intuition is that a path in the tree, which consists of several features, is a good candidate of a counting key. To achieve this goal, we independently learn one tree-based model per user for a selected set of users, who has sufficient impressions data. We use a frequency-based measure as weighting mechanism for each path in the tree. The paths with the highest weights are selected as our counting keys, which guarantees our counting features have good coverage.  Through extensive experiments with Twitter real ads traffic, we show that the selected counting features consistently improve the CTR prediction performance. The contributions of this paper are summarized as follows. First, we propose a tree-based method to automatically select counting keys. Second, we compare the automatically selected counting features with a set of human curated features, and show better performance. Third, we evaluate our method with two widely used setting, i.e. online learning with logistic regression and offline training with neural network models, and show consistent results.

The result of this paper is organized as follows. In Section~\ref{sec:rela-work}, we present related works in the literature. In Section~\ref{sec:prob-def}, the formal definition of our problem is given. The method is described in Section~\ref{sec:method}, and the experiment results are shown in Section~\ref{sec:exp}. Finally, we conclude our paper with future work in Section~\ref{sec:conclu}.

\section{Related Work}
\label{sec:rela-work}

\textbf{CTR Prediction} is a well-studied problem. Logistic regression model is first used for CTR prediction at scale~\cite{battelle2011search, richardson2007predicting}. Later, online learning is used in ads system~\cite{ciaramita2008online, graepel2010web}, due to the rapid growth of data volume. To highlight the relationship and differences between theoretical advance and practical engineering in CTR prediction, McMahan et. al. \cite{mcmahan2013ad} at Google shares insights and lessons learned from the large-scale online learning CTR prediction system. Facebook~\cite{he2014practical} employs a gradient boosted decision tree (GBDT) to learn a feature transformation, and shows that applying the logistic regression on the transformed features achieve better performance. Li et al.~\cite{li2015click} shows that a learning-to-rank model, which learns the relative order of ads, actually outperforms a model that independently learns the CTR of each impression with Twitter timeline advertising data. As the content of ads is getting rich with image and video, deep learning model are used to account for the visual information~\cite{chen2016deep}.

As we can see, feature engineering is crucial to improve CTR prediction. In this paper, we aim to achieve the same goal by constructing effective historical features.

\textbf{Personalization} is an import aspect in CTR prediction to improve the long-term user engagement~\cite{wu2017returning}. Various probabilistic models are proposed to model the latent factors on user domain~\cite{liu2010personalized,cheng2010personalized}. For example, Shen et al.~\cite{shen2012personalized} propose to use graphic model and tensor decomposition technique to learn the latent factor of users. In highly dynamic domains, such as news recommendation and digital advertisement, exploration-exploitation strategies are widely adopted for personalization. The multi-armed bandit is a typical algorithm of such strategy. Li et. al.~\cite{li2016collaborative} use multi-armed bandit to dynamically group similar users.

Note that in this paper, we devise a counting feature selection method with personalization as a constraint. We differ from the personalization literature in a way that we are not trying to innovate new personalization method.
I 

\textbf{Attribute Interactions.} Within a counting key, a group of features jointly act as a CTR indicator. A relevant concept is attribute interaction~\cite{freitas2001understanding}, which refers to the fact that two or more attributes jointly present strong correlation with label, while each of them loosely correlate with the label. Here the attribute and feature are used interchangeably. There are literatures focusing on the analysis of attribute interaction~\cite{jakulin2003analyzing}, test the significant of feature interaction~\cite{jakulin2004testing}, etc.

We should note that counting keys selection is more complicated than attribute interaction. The attribute interaction mainly studies the interactions of several features within one sample. Meanwhile, the counting features are constructed from several features and aggregated over multiple historical samples. 

\textbf{Feature Selection}.
Finally, we want to emphasize that the feature selection~\cite{molina2002feature,guyon2003introduction,liu2005toward,chandrashekar2014survey} is another highly relevant research field. Feature selection aims at finding a subset of features that are representative to original data. Various feature selection methods are proposed in the literature, and there are mainly three categories~\cite{chandrashekar2014survey}. (1) Filter-based method~\cite{blum1997selection} defines a relevance score for each feature and select features with certain threshold. (2) Wrapper methods use the predictor as a black box and the predictor performance as the objective function to evaluate the feature subset. Examples are Sequential Feature Selection method~\cite{reunanen2003overfitting} and Genetic Algorithm~\cite{goldberg1989genetic}. (3) Unsupervised feature selection methods~\cite{mitra2002unsupervised,li2012unsupervised} are proposed as well, where clustering is a primarily used technique.

These feature selection methods cannot be used in our problem, because they do not address our challenges.  In addition to the exponential search space, implementing and storing the counting features are costly as well. What's more, the conventional feature selection method does not consider the sparsity of features at all.

\section{Problem Definition}
\label{sec:prob-def}

\newcommand{\cf}{\mathbf{c}} % contextual feature
\newcommand{\hf}{\mathbf{h}} % historical / counting feature
\newcommand{\uf}{\mathbf{u}} % user feature
\newcommand{\af}{\mathbf{a}} % ad feature
\newcommand{\Ds}{\mathcal{D}}
\newcommand{\Us}{\mathcal{U}}
\newcommand{\Is}{\mathcal{I}}

Given an online advertising system, we have millions of users denoted as $\Us$ and tens of thousands of ads denoted as $\Ds$. The objective of an advertising system is to present the most relevant ads to users. CTR is the most commonly used measure for ads relevance, which measures the probability that a user clicks on an ad.

Usually there is one prediction model $f$ to be learned for thousands of millions of impressions. Each impression contains the information about showing ad $d$ to user $u$ within certain context. It is important to differentiate different users and ads. To this end, we extract as many contextual features as possible for each impression, denoted as $\cf$. The contextual feature vector $\cf = [ c_1, c_2, \cdots c_n ] \in \mathbb{R}^N$ is an $N$-dimension vector, and each dimension $c_j$ corresponds to one feature. There are three types of contextual features, which are user features, ad features, and context features. The user features, such as user gender and age, differentiate users. The ad features are used to differentiate ads, and examples are advertiser category and ads content. The context features track the context of current impression, including time of impression, ad injection location, and nearby organic Tweets information.

The ads relevance prediction problem is defined as follows.
\begin{definition}[CTR prediction]
\label{def:ctr}
Given a set of impressions $\Is$, each impression $i$ denotes one event that ad $d_i \in \Ds$ is shown to user $u_i \in \Us$. We extract contextual feature $\cf_i$ for impression $i$. The CTR prediction aims to learn a prediction model
\[ \hat{p}_i = f(\cf_i), \]
where $\hat{p}_i$ is the predicted CTR.
\end{definition}

Another important signal for CTR prediction is the user's past engagements, namely historical features, which is missing in Definition~\ref{def:ctr}. We use counting features as one implementation of the historical engagement. To give the formal definition of counting features, we first define counting keys.

\begin{definition}[Counting key]
\label{def:ck}
A counting key $k$ is a combination of two or more contextual features, denoted as a feature set $k = \{ c_{p1}, c_{p2}, c_{p3}, \cdots \}$. Without loss of generality, we assume $c_{pi}$ are discrete features. Each counting key $k$ is defined over a pair of user and ad $\langle u, d \rangle$.
\end{definition}

\begin{definition}[Counting feature]
Counting feature accounts for the past engagements over a pair of user and ad $\langle u, d \rangle$. Given a counting key $k$ we are able to generate several counting features for user-ad pair $\langle u, d \rangle$, such as number of past impressions as $h_k^i$, number of past engagements as $h_k^e$, and past CTR as $h_k^p$. For each impression $i$ and a given counting key set $\mathcal{K} = \{k_1, k_2, \cdots \}$, we extract all counting features as $\hf_i = \{h_{k1}^i, h_{k2}^i, \cdots, h_{k1}^e, h_{k2}^e, \cdots, h_{k1}^p, h_{k2}^p, \cdots \} $.
\end{definition}

In this paper, our goal is to automatically select counting keys.
\begin{definition}[Counting feature keys selection]
Counting feature keys selection problem aims to find a set of counting keys $\mathcal{K}$ from given contextual feature $\cf$. These  is to generate counting features $\hf$ for each impression, and to improve the CTR prediction model through
\[ \hat{p}_i = f(\cf_i, \hf_i). \]
\end{definition}
\section{Method}
\label{sec:method}

\subsection{Overview}

\begin{figure}[t]
\centering
\includegraphics[width=0.9\linewidth]{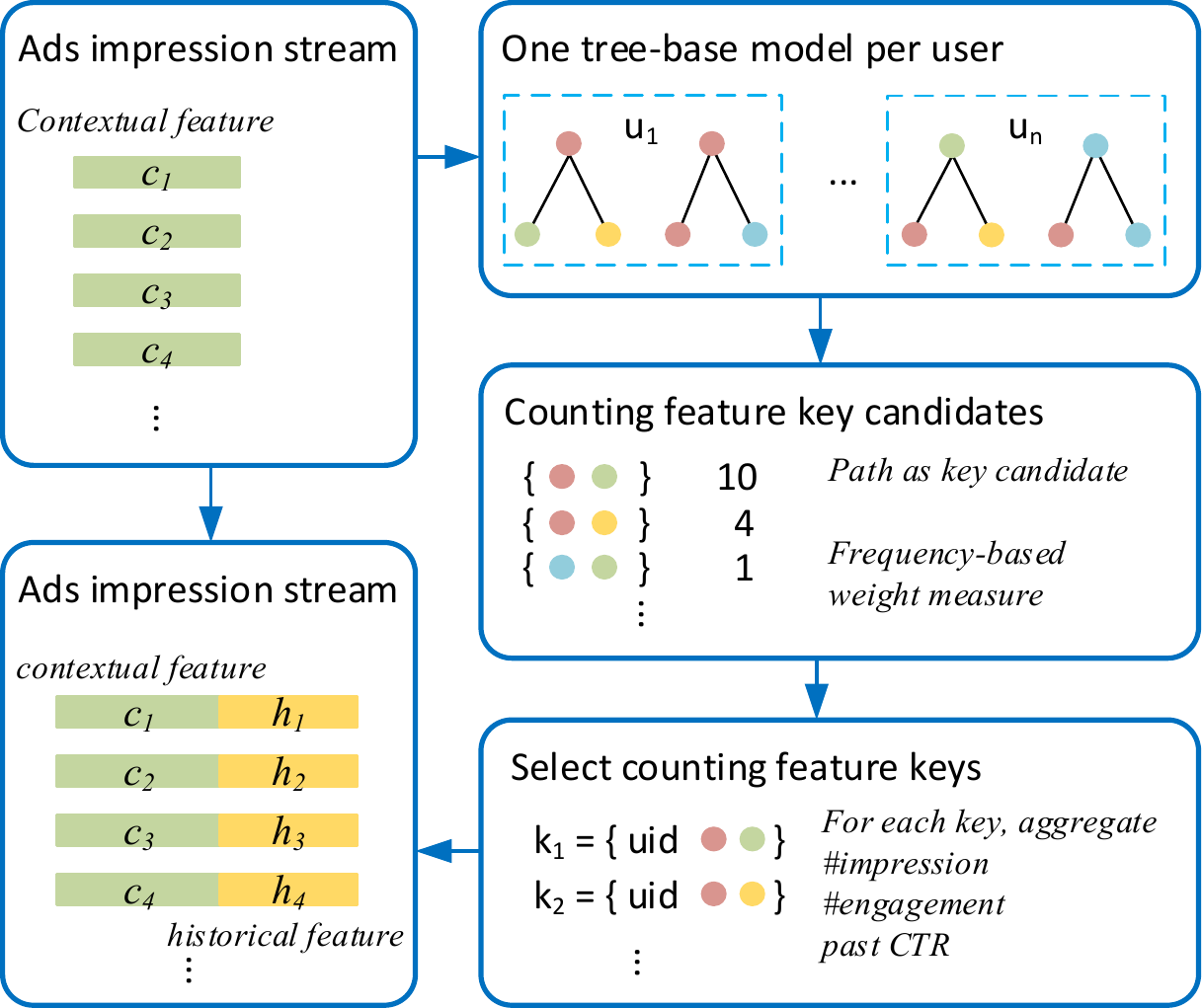}
\caption{The flow of automatic counting key selection. We use $c_i$ and $h_i$ denote contextual features and historical features for impression $i$, respectively. Symbol $u_i$, $i\in \{1, 2, \cdots, n\}$ refers to user $i$, and $uid$ is short for user ID. Each colored circle is one type of feature. And $k_i$ is one counting key, which is a set of features including user id.}
\label{fig:overview}
\end{figure}

We visualize the overview flow of our method in Figure~\ref{fig:overview}. We take a stream of impressions as input, where the contextual features (green features denoted as $\cf_i$) of impressions are available. Next, in order to generate the counting key candidate, we learn one tree-based binary classification model per user to predict the click label. The motivation behind training one tree per user is personalization. A simplest approach to achieve personalization in the counting features is to keep the user ID as one component in the key. With the tree-based model, each path in the trees is a set of features that together act as strong indicator for click event. The counting key candidate is a set of features, which is constructed by adding user ID into the feature set from one tree path. In the end, we have many counting key candidates from those tree-based models. In order to find counting keys that are discriminative to all users, we use a frequency-based measure as the weighting measure. Finally, we generate counting features for each pair of user and ads according to the selected counting keys. These counting features (yellow features denoted as $\hf_i$) are joined with the contextual features to serve the CTR prediction.

\subsection{Counting Key Candidates}

We use decision trees to generate counting key candidates. In order to implement personalization, we therefore build one decision tree model $M_i$ for user $u_i$. The input for decision model $M_i$ consists of all impressions related to user $u_i$. 

In our system, we use Xgboost~\cite{chen2016xgboost} to build one gradient boosted forest for each user. In gradient boosted forest, trees are built in sequential order, where the newly built trees aim to minimize the errors from previous trees. The oldest trees usually capture the most important features to differentiate general cases, while the latest trees usually account for the specific features to differentiate the difficult cases. With these boosted decision trees, we can use all the paths as counting key candidates.

\subsubsection{Theoretical Support}

Now we provide theoretical support to explain why tree-based model could be used to select counting keys. The objective of counting key selection is to identify a set of features $k = \{c_{p1}, c_{p2}, \cdots \}$, that are the discriminative for click prediction. We use the conditional entropy to measure the quality of a counting key $k$. We use $Y$ to denote the click label. If a feature key $k$ is useful, then the conditional entropy of $H(Y|k)$ should be less than the marginal entropy $H(Y)$, i.e. $H(Y|k) < H(Y)$. In order to find the best counting key, we maximize the difference between the two. Formally, we have the following objective:
\begin{equation}
\label{eq:obj}
\arg\max_k H(Y) - H(Y|k).
\end{equation}
The conditional entropy $H(Y|k)$ is defined as 
\begin{equation}
\label{eq:cond-entro}
H(Y|k) = \sum_{ v \in vals(k) } p(v) H(Y|k = v),
\end{equation}
where $vals(k)$ refers to all possible values that vector $k$ could take.

The Equation~(\ref{eq:obj}) above is difficult to optimize, since the $k$ is a set of features with exponential search space. Also, the goal of counting key selection is not to find the best counting key, but rather to find as many counting keys as possible. Those found counting keys should satisfy the condition that $H(Y) - H(Y|k) > \Delta$, where $\Delta$ is a threshold on entropy decrease.

The tree-based model selects feature and split value to maximize the information gain. We note that the information gain in tree-based model is calculated on a binary split, which is not exactly equivalent to conditional entropy. However, we prove that information gain of a binary split is a lower-bound of $H(Y) - H(Y|k)$.

\begin{figure}[t]
\begin{minipage}[t]{0.4\linewidth}
\centering
\includegraphics[width=0.8\linewidth]{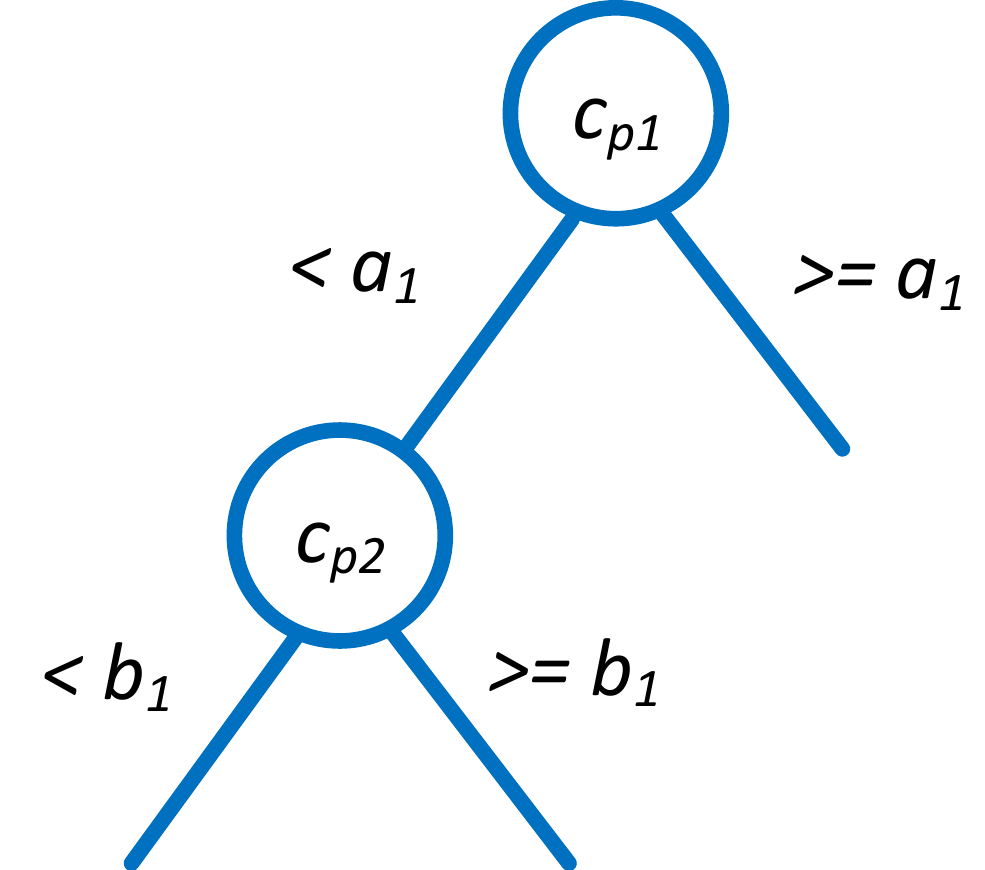}
\end{minipage}
\begin{minipage}[t]{0.4\linewidth}
\centering
\begin{tabular}{|c|c|c|c|}
\hline
Feature & \multicolumn{3}{|c|}{Value} \\ \hline
$c_{p1}$ & $a_0$ & $a_1$ & $a_2$ \\ \hline
\end{tabular}
\end{minipage}
\caption{Tree-based model example. Suppose the feature $c_{p1}$ only takes three values $a_0$, $a_1$, and $a_2$, where $a_0 < a_1 < a_2$.}
\label{fig:tree}
\end{figure}

Without loss of generality, we consider the tree-based model in Figure~\ref{fig:tree}. The first split maximizes the information gain $IG(Y, c_{p1}, a_1)$, which means split the samples with feature $c_{p1} < a_1$.
\begin{multline}
\label{eq:ig}
 IG(Y, c_{p1}, a_1) =  H(Y) - \Big( P(c_{p1} < a_1) H(Y|c_{p1} < a_1) \\
 + P(c_{p1} \ge a_1) H(Y|c_{p1} \ge a_1) \Big)
\end{multline}

Apply the inequality property of conditional entropy, we have 
\begin{multline}
\label{eq:deduction}
H(Y|c_{p1} \ge a_1) \ge P(c_{p1} = a_1 | c_{p1} \ge a_1) H(Y|c_{p1} = a_1) \\ 
+ P(c_{p1} = a_2| c_{p1} \ge a_1) H(Y|c_{p1} = a_2)
\end{multline}

Take Equation~(\ref{eq:deduction}) into Equation~(\ref{eq:ig}), and notice that $c_{p1} < a_1 \iff c_{p1} = a_0$. We have
\begin{multline}
\label{eq:ig}
 IG(Y, c_{p1}, a_1) \le  H(Y) - \Big( P(c_{p1} = a_0) H(Y|c_{p1} = a_0) \\
 + P(c_{p1} \ge a_1) P(c_{p1} = a_1 | c_{p1} \ge a_1) H(Y|c_{p1} = a_1) \\
 + P(c_{p1} \ge a_1) P(c_{p1} = a_2| c_{p1} \ge a_1) H(Y|c_{p1} = a_2) \Big).
\end{multline}

Therefore, we prove that the information gain is a lower-bound of $H(Y) - H(Y|c_{p1})$, i.e.
\begin{equation}
\label{eq:approximation}
IG(Y, c_{p1}, a_1) \le H(Y) - H(Y|c_{p1}).
\end{equation}

Follow the left path of tree, the model recursively maximizes the $IG(Y, [c_{p1}, c_{p2}], [a_1,b_1])$. By applying Equation~(\ref{eq:approximation}), we have that the $IG(Y, [c_{p1}, c_{p2}], [a_1,b_1]) \le H(Y) - H(Y|c_{p1}, c_{p2})$. Namely, the tree-based model maximizes the lower-bound of $H(Y) - H(Y|k)$.

\subsection{Counting Key Quality Measure}

Although we can limit the number and depth of trees, there are still too many counting key candidates. We should also notice that the latest trees usually contain counting keys that are very specific for one user, but not appear in other users' models. Meanwhile, it is possible that the oldest trees provide us counting keys that are too general to be useful. A quantitative measure on the relevance of each counting key candidate is needed.

Motivated by the intuition above, we propose to use the term frequency and inverse document frequency (tf-idf)~\cite{ramos2003using} to measure the relevance of counting key candidates. We treat each user as one ``document'', and the feature set of each path as a ``word''. Notice that we omit the splitting value for each node, as well as the order of features in the path. The reason is that the counting feature key is a set, which does not differentiate the order of features in the key.

We use $\mathcal{M} = \{ M_i \}$ to denote all tree-based models. We use $\mathcal{M}_k = \{ M_i | k \models M_i \}$ to denote the set of models that contain candidate key set $k$. The notation $k \models M_i$  means there is a path in model $M_i$ such that the feature set on the path is $k$.

The term frequency (tf) is defined as
\begin{equation*}
\text{tf}(k, \mathcal{M} ) = \frac{ \sum_{j \in \mathcal{M}_k} f_j }{|\mathcal{M}_k|},
\end{equation*}
where $|\cdot|$ is the cardinality of a set, and $f_j$ is the number of times $k$ appears in model $M_j \in \mathcal{M}_k$.
The inverse document frequency (idf) is defined as 
\begin{equation*}
\text{idf}(k, \mathcal{M}) =\log \frac{ |\mathcal{M}| }{ 1 + |\mathcal{M}_k| }.
\end{equation*}

Finally, our counting key weight measure is
\begin{equation}
\label{eq:tfidf}
\text{tf-idf} = tf(k, \mathcal{M}) \cdot \text{idf}(k, \mathcal{M})
\end{equation}

\subsection{Apply Counting Features}

With the tf-idf measure, we pick the top counting key candidates as $\mathcal{K}$ to generate counting features. For each pair of user $u$ and ad $d$, we count three values on a given key $k$. They are number of impressions, number of engagements, and past CTR. Namely, we generate the historical features 
\[ \hf = \bigcup_{\forall k \in \mathcal{K}} \{ c_{k}^i, c_k^e, c_k^p \}. \]
As shown in the bottom left of Figure~\ref{fig:overview}, the historical features $\hf$ are concatenated with contextual features $\cf$ for the impression data. Together, they are used for the CTR prediction.

\section{Experiments}
\label{sec:exp}

\newcommand{\base}{\emph{BASE}\xspace}
\newcommand{\human}{\emph{HUMAN}\xspace}
\newcommand{\acf}{\emph{ACF}\xspace}

\newcommand{\lr}{\emph{LR}\xspace}
\newcommand{\wdnn}{\emph{NN}\xspace}

\subsection{Experiment Settings}

We use Twitter video advertisement data stream to evaluate our method. More specifically, we use data from 18 continuous days in 2017. The first 15-day data are used to select the counting feature keys and generate corresponding counting features for all users. Then, we use the data of the last 3 days to evaluate the automatically identified counting features. 

We compare three different feature settings to validate the effectiveness of automatically identified counting features.
\begin{enumerate}[leftmargin=*]
\item Without user-ads engagement features (\base), where we use only contextual features $\cf$ to predict the CTR.
\item Human curated user-ads engagement features (\human). These features, denoted as $\hf_h$ are generated from human heuristics and past experiences. In this setting, we use $\cf$ and $\hf_h$ together to predict CTR.
\item Automatically selected counting features (\acf). Our method automatically extracts a set of counting features, denoted $\hf_a$. We use $\cf$ and $\hf_a$ together to predict the CTR.
\end{enumerate}

With the three feature combinations, we show the CTR prediction results under two different model. 
\begin{enumerate}[leftmargin=*]
\item An online learning model using logistic regression (\lr). The ads engagement data is generated in a stream, and an online learning model constantly updates itself given new engagement data. A data sample (user-ad pair) first goes through a discretizer, which is a gradient boosted decision trees, to get compact representations. Then the compact vectors are fed into a logistic regression model to get the predicted CTR.
\item An offline training model using wide and deep neural network~\cite{cheng2016wide} (\wdnn). The wide component has a single layer, which acts as a linear model on the given raw features. The deep component consists of multiple layers, which aims to automatically learn the interactions among features. Two components are jointly learned and the final output is the predicted CTR.
\end{enumerate}

We use the relative cross entropy (RCE)~\cite{he2014practical} as our evaluation measure, mainly because we use cross entropy as our loss function,
\begin{equation}
\label{eq:ce}
CE = \sum_i \Big(- y_i \log \hat{p_i} - (1-y_i) \log (1-\hat{p_i}) \Big), 
\end{equation}
where $y_i \in \{0, 1\}$ is the label of each input and $\hat{p}$ is the predicted CTR.

The relative cross entropy is equivalent to the cross entropy of our model normalize by the cross entropy of a baseline model, which uses the background average CTR $p$ to predict every impression. Suppose the background average CTR is p, then the RCE is
\begin{equation}
\label{eq:rce}
RCE = \frac{\sum_i \big(- y_i \log \hat{p_i} - (1-y_i) \log (1-\hat{p_i}) \big)}{-p\log p - (1-p)\log (1-p)}.
\end{equation}
%The reason to use RCE instead of CE is that the log loss (CE) is sensitive to background CTR $p$. When $p$ is close to 0 or 1, it is easier to achieve a better log loss. The normalization in RCE solves this problem.

\subsection{Counting Features}
\label{sec:counting-features}

\subsubsection{Automatically Selected Counting Features -- \acf}
To obtain the counting feature candidates, we use the data for the first five days. Due to data sparsity, we only pick the top 100 users with the most impressions. We train one gradient boosted decision trees for each user. The number of trees is set to 20, and the maximum depth of each tree is set to 3. Here we notice that the size of a counting keys should not be too large, otherwise the counting value for such key will be too sparse. Considering that user ID is one feature in the counting keys, we set the maximum tree depth as 3.

We pick the following counting feature keys, shown in Table~\ref{tab:cfk}, according to their corresponding tf-idf values. Given the counting feature keys, we count the number of impressions, number of engagements, and the corresponding average CTR as features. In our experiments, we generate the values for the counting features using the 15-day data. We assume the counting features won't change in a short time. Therefore, we join these historical counting features with the next 3-day data for evaluation.

\begin{table}[h]
\centering
\caption{Automatically Extracted Counting Feature Keys}
\label{tab:cfk}
\begin{tabular}{|l|l|l|r|}
\hline
\multicolumn{3}{|c|}{Counting Keys} & tf-idf \\ \hline
\multirow{6}{*}{user ID} 
 	& item objective & - & 9.3 \\ \cline{2-4}
    & engagement option & - & 8.6\\ \cline{2-4}
	& hour of day & - & 4.6 \\ \cline{2-4}
    & hour of day & item objective & 4.4 \\ \cline{2-4}
    & hour of day & engagement option & 3.9 \\ \hline
\end{tabular}
\end{table}

\subsubsection{Human Curated Counting Features -- \human}
The \human features include the past CTR, number of impressions, and number of engagements on a set of human selected counting keys. These counting keys are selected based on human heuristics, where the selected features are all categorical. Some examples of selected features are advertiser ID, user age bucket, and device type. The total number of \human features is $453$. We should note that many of the \human counting features are very sparse.

%\begin{table}[h]
%\centering
%\caption{Features used in the \human counting key. Note the list is not exhaustive.}
%\label{tab:human-example}
%\begin{tabular}{|l|l|}
%\hline
%Feature & Description \\ \hline
%ad ID & Unique identifier for ads. \\ \hline
%advertiser ID & Unique identifier for advertisers. \\ \hline
%similar advertisers & Cluster of advertisers. \\ \hline
%age bucket & Discrete age of user. \\ \hline
%device type & Android, iOS, or desktop browser. \\ \hline
%\end{tabular}
%\end{table}

\subsection{CTR Prediction with Online Learning}

In this section we show the CTR prediction results in an online learning setting. This CTR prediction model consists of a gradient boosted decision tree and a logistic regression. The gradient boosted decision tree (GBDT) is used generate new features, and it is learned independently from the latter logistic regression. The goal of GBDT is to deal with the feature non-linearity. The logistic regression is subsequently applied on the binary vectors from GBDT. In an online learning setting, we update the logistic regression model as new data coming in. In total, 1\% data are sampled as testing data, and the rest data are used as training. Impression are fed to the model in chronological order. The testing samples are not used to update the model. When encounter a testing sample, we use current model to report predicted CTR, which is later used to calculate the testing RCE.

%\begin{table}[h]
%\centering
%\caption{The average online learning RCE of various feature settings on last day.}
%\label{tab:rce-lr}
%\begin{tabular}{|c|r|}
%\hline
%Feature Settings & RCE \\ \hline
%\base & 12.874 \\ \hline
%\human & 12.825 \\ \hline
%\acf & 12.952 \\ \hline
%\end{tabular}
%\end{table}

\begin{figure}[t]
\centering
\includegraphics[width=0.7\linewidth]{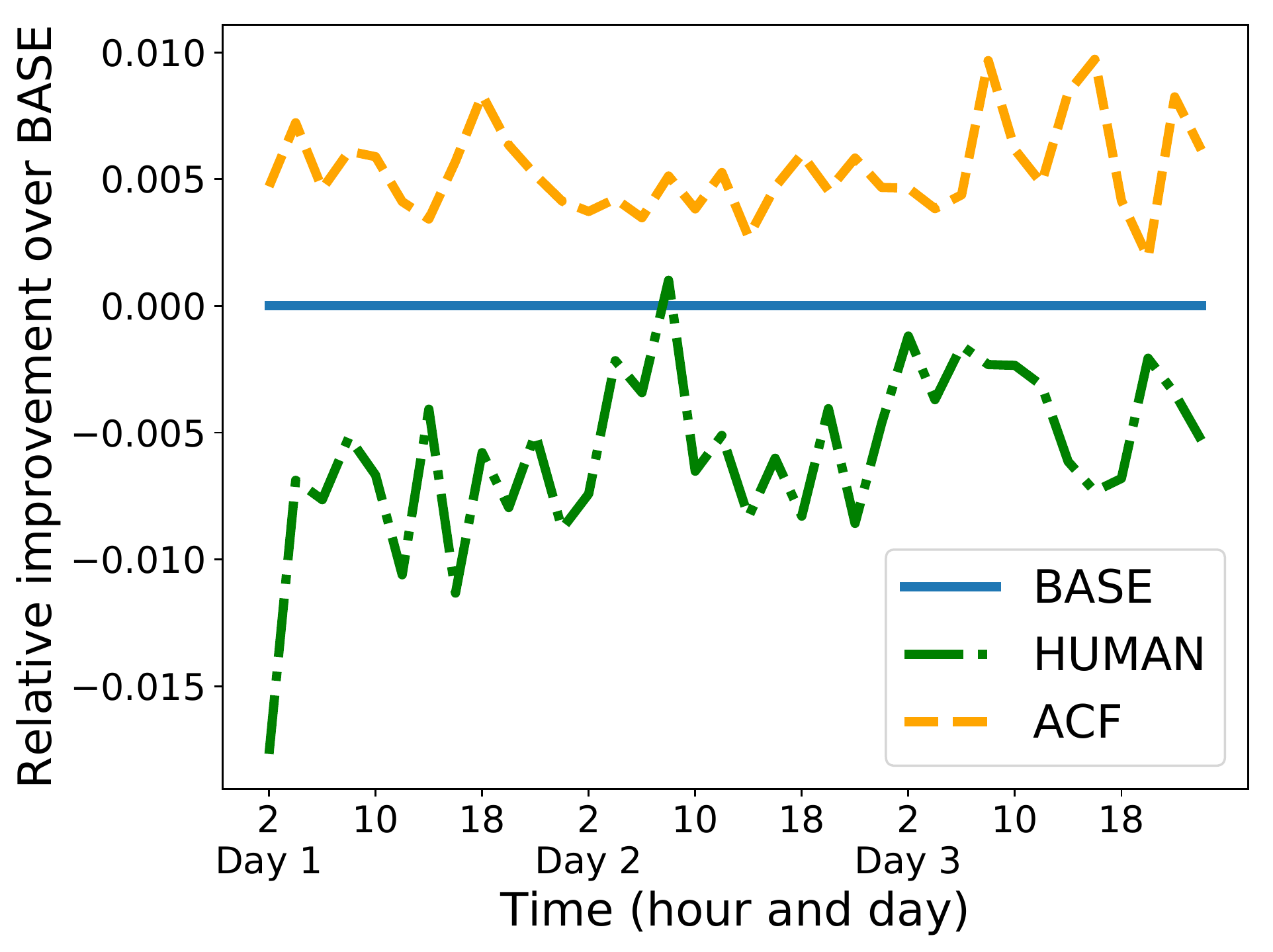}
\caption{The relative RCE improvement over BASE method. Evaluated with logistic regression model in an online learning setting for different features.}
\label{fig:rce}
\end{figure}

Considering the model warms up need some time, we report the average RCE on the last day. Our \acf features has a $0.6\%$ higher RCE than the \base features, and the \human features actually are worse than the \base method. Although such RCE gain seems small, the revenue impact is still huge at scale. Also, we should keep in mind that \acf only contains a dozen features, while the quantity of \human is close to $500$. In order to look into the model warming-up process, we further plot the RCE curve over time in Figure~\ref{fig:rce}.

In Figure~\ref{fig:rce}, we report the relative RCE improvement over \base method over three days at two-hour interval. It is clear that our \acf features consistently outperforms the \base features. Meanwhile, we notice that \human features are almost consistently worse than \base features. We further investigate this issue in Section~\ref{sec:feat-quality}, and find the main reason is that \human contains too many sparse features, which hurts the model performance.

\subsection{CTR Prediction with Batch Training}

\begin{figure}[t]
\centering
\includegraphics[width=0.8\linewidth]{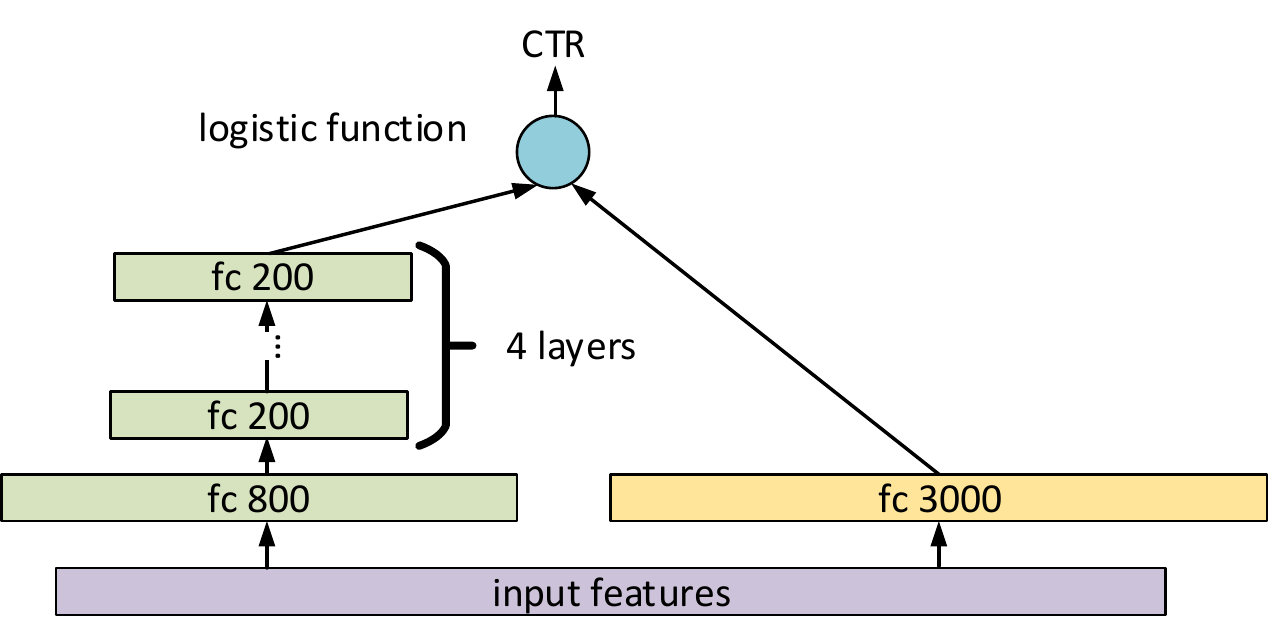}
\caption{Wide \& Deep \wdnn model structure.}
\label{fig:nn-struct}
\end{figure}

In this section we show the CTR prediction results in a batch training setting. We evaluate our new features in a batch training setting with a wide and deep \wdnn model shown in Figure~\ref{fig:nn-struct}. More specifically, we use the first 2-day data as training data for the \wdnn model, and test on the last day. 

%The \wdnn model combines the generalization power of a neural network and the memorization power of a linear model. As shown in Figure~\ref{fig:nn-struct}, the same impression features are fed into both components. On the left, a 5-layer neural network learns the combination of various features. On the right, a linear model act as the logistic regression to learn weight on each of the original features. 

The \acf achieves the best performance, with a testing RCE $2.02\%$ better than \base. And \human outperforms the \base by $1.06\%$ in RCE, mainly because the 5-layer fully connected model could learn new combinations of features.

%\begin{table}
%[h]
%\centering
%\caption{The offline training RCE for various feature settings. Trained on first two days. Tested on last day.}
%\label{tab:rce-nn}
%\begin{tabular}{|c|r|}
%\hline
%Feature Settings & RCE \\ \hline
%\base & 9.3693 \\ \hline
%\human & 9.4960 \\ \hline
%\acf & 9.5582 \\ \hline
%\end{tabular}
%\end{table}

\subsection{Feature Analysis}
\label{sec:feat-quality}

In this section, we compare the feature quality of \acf and \human. 

\textbf{Feature Coverage} We define the feature coverage as the percentage of number of samples that has non-zero value on given feature. We summarize the key coverage statistics for \human and \acf in Table~\ref{tab:cf-cov-comp}. Notice that there are \human features that cover no impressions, because they are too sparse.

\begin{table}[h]
\centering
\caption{Feature coverage comparison.}
\label{tab:cf-cov-comp}
\begin{tabular}{|c|c|c|}
\hline
Feature type & \acf & \human \\ \hline
Number of features & 15 & 453 \\ \hline
Average coverage & 0.8411 & 0.0286 \\ \hline
Maximum coverage & 0.9768 & 0.9461 \\ \hline
Minimum coverage & 0.6179 & 0 \\ \hline
\end{tabular}
\end{table}

\textbf{Feature Correlations.} We calculate the Pearson correlation between each feature and click label. For each feature, since the coverage is not $100\%$, we only calculate the correlation on a subset of impressions where the feature values are not missing. A summary of correlations of both features are compared in Table~\ref{tab:cf-cor-comp}, from which it is easy to tell that overall correlations of \acf are better than \human ($0.0451$ vs. $0.0138$). 

\begin{table}[h]
\centering
\caption{Feature correlation comparison.}
\label{tab:cf-cor-comp}
\begin{tabular}{|c|c|c|}
\hline
Feature type & \acf & \human \\ \hline
Number of features & 15 & 102 \\ \hline
Average correlation & 0.0451 & -0.0003 \\ \hline
Maximum correlation & 0.2026 & 0.0138 \\ \hline
Minimum correlation & 0.0005 & -0.0165 \\ \hline
\end{tabular}
\end{table}

\section{Conclusion}
\label{sec:conclu}

In this paper, we study how to automatically generate counting features in ads CTR prediction. We propose a tree-based method to automatically select counting feature keys. The proposed method is simple and efficient. Through extensive experiments with Twitter real data, we validate that the automatically extracted counting features have better coverage and higher correlations with the click labels, compared to human curated features. It is noteworthy that our approach could be easily generalized beyond user-centric counting features, such as advertiser-centric and interest-centric counting features.

\bibliographystyle{ACM-Reference-Format}
\bibliography{reference} 

\end{document}